\documentclass[conference]{IEEEtran}
\IEEEoverridecommandlockouts
\pdfoutput=1
\usepackage{cite}
\usepackage{amsmath,amssymb,amsfonts}
\usepackage{algorithmic}
\usepackage{graphicx}
\usepackage{textcomp}
\usepackage{xcolor}
\usepackage{tikz}
\usepackage{graphicx}
\usepackage{float}
\usetikzlibrary{positioning}

\def\BibTeX{{\rm B\kern-.05em{\sc i\kern-.025em b}\kern-.08em
    T\kern-.1667em\lower.7ex\hbox{E}\kern-.125emX}}

\begin{document}

\title{Transfer Learning using Neural Ordinary Differential Equations\\
}

\author
{
\IEEEauthorblockN{Rajath S}
\IEEEauthorblockA{\textit{Student,} \\ 
\textit{Department Of Computer Science} \\
\textit{PES University}\\
Bengaluru,India \\
rajaths2510@gmail.com}
\and
\IEEEauthorblockN{Sumukh Aithal K}
\IEEEauthorblockA{\textit{Student,} \\ 
\textit{Department Of Computer Science} \\
\textit{PES University}\\
Bengaluru,India \\
sumukhaithal6@gmail.com}
\and
\IEEEauthorblockN{Dr.S Natarajan}
\IEEEauthorblockA{\textit{Professor,}\\ \textit{Department Of Computer Science} \\
\textit{PES University}\\
Bengaluru,India \\
natarajan@pes.edu}

}

\maketitle

\begin{abstract}
We introduce a concept of using Neural Ordinary Differential Equations(NODE) for Transfer Learning. In this paper we use the EfficientNets to explore transfer learning on CIFAR-10 dataset. We use NODE for fine-tuning our model. Using NODE for fine tuning provides more stability during training and validation.These continuous depth blocks can also have a trade off between numerical precision and speed .We conclude that the using Neural ODEs for transfer learning results in much stable convergence of the loss function.                                                
\end{abstract}


\section{Introduction}
Image classification is one of the fundamental tasks in computer vision and there has been significant improvement in the accuracy of image classification models since CNNs. AlexNet \cite{krizhevsky2012imagenet} and GoogleNet  \cite{szegedy2015going} showed that deeper and larger neural network models perform better at image classification. CNNs learn by feature extraction and features extracted by one model trained on a particular dataset can be used by another model performing a similar task. \newline

Given the enormous amount of resources used to train computer vision models,transfer learning is a very popular technique used in deep learning.Transfer learning significantly improves the training time and gives better results compared to the conventional techniques.
While most machine learning algorithms are designed to address single tasks, the development of algorithms that facilitate transfer learning is a topic of ongoing interest in the machine-learning community. 
\newline

In this paper we study the concept of using NODE for transfer learning by using EfficientNet model as the backbone and imagenet weights as the pretrained weights.The intuition behind it being that brain is also considered as continuous time systems.

The remaining structure of the paper is as follows.Section II contains the related work.The proposed method and details of the experiments is explained in Section III.Results and Conclusions is explained in Section IV and V respectively.
\section{Related Work}

EfficientNets \cite{tan2019efficientnet} are a family of models which can be systematically scaled up based on the resources available.This family of models achieve better accuracy than traditional ConvNets.
In these models , there is a principled way in which models are scaled up . A balance between width,depth and height is achieved by simply scaling them up with a constant ratio.
These models can be scaled up based on the computation resources available.For instance if there exists $2^n$ more resources then the depth could be increased by $ \alpha^n$, width by $\beta^n$ and image size by $\gamma^n$ where $\alpha, \beta,\gamma$
are constant coefficients
EfficientNets transfer well on datasets like CIFAR-100 \cite{krizhevsky2009learning}, Flowers etc with fewer parameters.
Tan et al,\cite{tan2019efficientnet}, have examined eight models from EfficientNetB0-EfficientNetB7 for their efficiency and performance.
\bigbreak

 Neural Ordinary Differential Equations (NODE)\cite{chen2018neural} are a family of Neural Networks where discrete sequence of hidden layers need not be specified, instead the derivative of the hidden state is parameterized using a neural network.Networks such as ResNets\cite{he2016deep} and Recurrent Neural Networks\cite{mikolov2010recurrent} can be modelled as continuous transforms using NODE.
 These continuous-depth models have constant
memory cost, adapt their evaluation strategy to each input, and can explicitly trade
numerical precision for speed.\newline

 Chen et al.,\cite{chen2018neural} use adaptive step-size solvers to solve ODEs reliably. This solver uses the adjoint method\cite{melicher2017fast} and the network used has a memory cost of O(1).
 Here, a network is also tested with the same architecture but where gradients are backpropagated
directly through a Runge-Kutta integrator, referred to as RK-Net, which has O(L) memory cost where L stands for the number of layers in the network.
Here , the continuous
dynamics of hidden units are parameterized using an ordinary differential equation (ODE) specified by a neural network:
\begin{equation}
\frac{dh(t)}{dt} = f(h(t),t,\theta) 
\end{equation}
where $t \in {0 . . . T}$ Input layer would be h(0), output layer can be defined as h(T) to be the solution to this
ODE initial value problem at some time T.\newline \\
Currently, as numerical instability is an issue with ODEs,
augmented version of Neural ODE networks have been developed\cite{dupont2019augmented}\cite{zhang2019anodev2}.
Among many extensions of Neural ODEs developed, one such is an approach that  allows evolution of the neural network parameters, in a coupled ODE-based formulation. Also Augmented Neural ODEs  are modelled which, in addition to being more expressive models than traditional Neueral ODEs, are empirically more stable, generalize better and have a lower computational cost than Neural ODEs
\cite{zhang2019anodev2}. \\
\\
Having stability while training deep Neural Networks is important as
it  consistently offers improved robustness against
a broader range of distortion strengths and types unseen during training,
a considerably smaller hyperparameter dependence and less potentially
negative side effects compared to data augmentation\cite{laermann2019achieving}. \\
Stability during training is also achieved using different activation functions such as bounded Rectified Linear Unit (ReLU), bounded leaky ReLU, and bounded bi-firing\cite{liew2016bounded}.\\
\\
There are many domains where well annotated data is not easy to obtain due to data acquisition expenses. Collection of data is complex and expensive that make it extremely difficult
to build a large-scale, high-quality annotated dataset. Transfer learning is the solution to this problem as it relaxes the hypothesis that the training data must be independent and identically distributed with the test data \cite{tan2018survey}.

\section{Proposed Method}
In the proposed method  EfficientNet B0 is used, from the family of EfficientNet models as the base model. A NODE layer is added before the final layer for fine tuning the CIFAR-10 dataset\cite{krizhevsky2009learning}.\newline \\ Although it increases the time taken to train per epoch,the NODE block is added to gain stability in this process.
Two ODE Solvers for NODE i.e Runge Kutta Method\cite{schober2014probabilistic} and the modern adjoint sensitivity method\cite{pontryagin2018mathematical} are used and compared. 
The default implementation tf.contrib.integrate.odeint is used to solve the Ordinary Differential Equation initial value problems . This default method uses the  Runge-Kutta solver.  Adjoint sensitivity method is also used to for better memory efficiency.
The adjoint method is computationally efficient and is numerically much more stable.
\newline \\
While using the Runge-Kutta(RK) Method, a user defined hyperparameter, relative and absolute tolerance limit can be varied to achieve optimal performance.Setting both these parameters as \(10^{-5}\) provided optimal results .Tolerance limit is a parameter which is a trade off between accuracy and computational cost. \newline
For the Adjoint sensitivity method, the default parameters provided by Chen et.al. \cite{chen2018neural} is used.\\
\newline
The proposed model consists of two variations, one with RK solver which is run for 200 epochs and one with the modern adjoint sensitivity method which is run for 160 epochs.\\
In the proposed method,  EfficientB0 model is run for 200 epochs to obtain the desired accuracy as mentioned in the performance table .The NODE block is then added to the previous model and is run until the desired validation accuracy is observed. It is then observed that the same validation accuracy  obtained after 100 epochs in the case of RK solver and 160 epochs in the case of the adjoint sensitivity method.\\
\newline
For the initial model without NODE and the proposed model with RK solver, the Adam optimizer\cite{kingma2014adam} was used. For the proposed model with the Adjoint method,  stochastic gradient descent\cite{bottou2010large} was better than the Adam Optimizer.\\
Using ImageNet\cite{deng2009imagenet} weights for training the model enables quicker convergence as the features learnt by the pre-trained model are common to the image classification task.
\newline
The concept of using NODE instead of the fully connected layer is that it potentially reduces the number of parameters, as the hidden blocks are now continuous functions of time. Also one can tune the tolerance parameter for speed/accuracy trade-off.

\begin{figure}[ht]

  \centering
  \includegraphics[width=.3\linewidth,height=8cm]{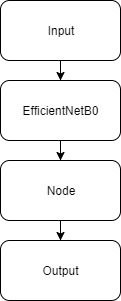}
  \caption{Model Visualization}
  \label{}

\end{figure}
\section{Results and Conclusion}
\begin{figure}[!h]
    \centering
    \includegraphics[scale=0.7]{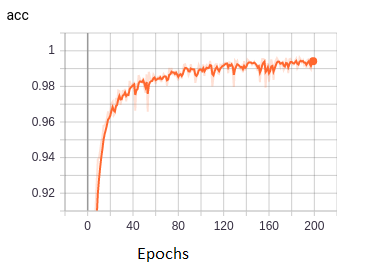}
    \caption{Training accuracy}
    \label{fig:fig2}
\end{figure}

\begin{figure}[!h]
    \centering
    \includegraphics[scale=0.7]{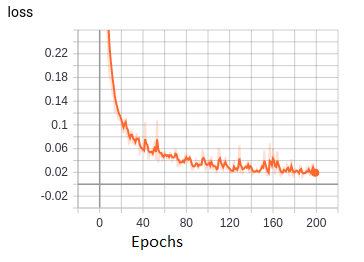}
    \caption{Training loss}
    \label{fig:fig3}
\end{figure}

\begin{figure}[!h]
    \centering
    \includegraphics[scale=0.7]{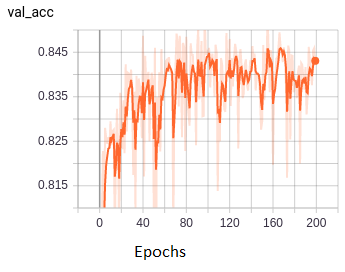}
    \caption{Validation accuracy}
    \label{fig:fig4}
\end{figure}

\begin{figure}[!h]
    \centering
    \includegraphics[scale=0.7]{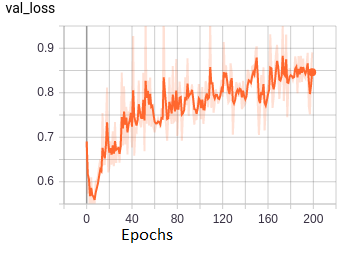}
    \caption{Validation loss}
    \label{fig:fig5}
\end{figure}

\begin{figure}[!h]
    \centering
    \includegraphics[scale=0.7]{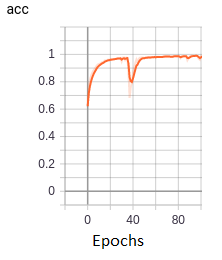}
    \caption{Training Accuracy(proposed model with RK solver)}
    \label{fig:fig6}
\end{figure}

\begin{figure}[!h]
    \centering
    \includegraphics[scale=0.7]{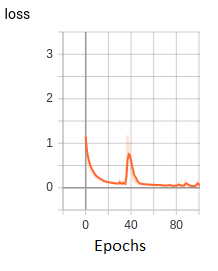}
    \caption{Training loss(proposed model RK solver)}
    \label{fig:fig7}
\end{figure}

\begin{figure}[!h]
    \centering
    \includegraphics[scale=0.7]{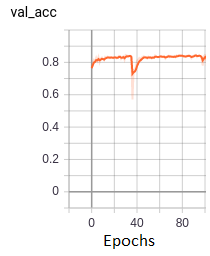}
    \caption{Validation Accuracy(proposed model RK solver)}
    \label{fig:fig8}
\end{figure}

\begin{figure}[!h]
    \centering
    \includegraphics[scale=0.7]{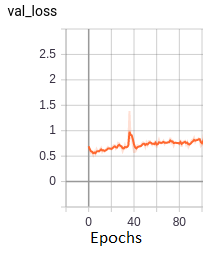}
    \caption{Validation loss(proposed model RK solver)}
    \label{fig:fig9}
\end{figure}

\begin{figure}[!h]
    \centering
    \includegraphics[scale=0.7]{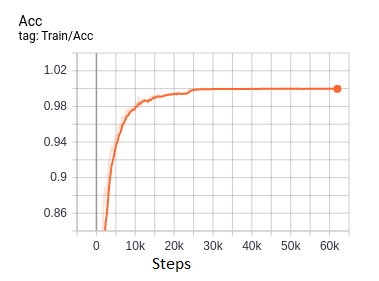}
    \caption{Training accuracy using adjoint solver(proposed model)}
    \label{fig:fig10}
\end{figure}

\begin{figure}[!h]
    \centering
    \includegraphics[scale=0.7]{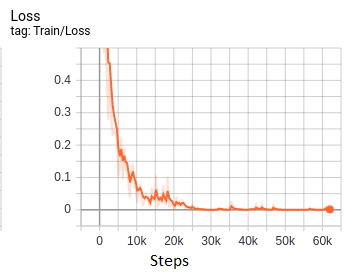}
    \caption{Training loss using adjoint solver(proposed model)}
    \label{fig:fig11}
\end{figure}

\begin{figure}[!h]
    \centering
    \includegraphics[scale=0.7]{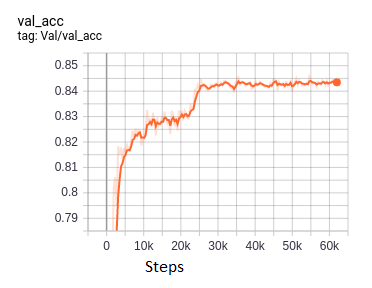}
    \caption{Validation Accuracy using adjoint solver(proposed model)}
    \label{fig:fig12}
\end{figure}

It is observed that using NODE  before the final layer guarantees better stability during training and validation.  Both results are shown in Table \ref{table:table3}, the one with NODE at the end and the one without NODE(purely EfficientNetB0 with just a final fully connected layer). Results of both the variations of the proposed model is shown.\newline The model with EfficientNetB0 and a final fully connected layer is trained for 200 epochs.\newline The proposed model with RK solver is trained for 100 epochs while the model with Adjoint solver is trained for 160 epochs. It is observed that in both variants, the proposed model converges to the desired accuracy and loss much quicker than the model without NODE.The performance and stability get enhanced in the proposed model.\newline \\ The first four figures-\ref{fig:fig2},\ref{fig:fig3},\ref{fig:fig4},\ref{fig:fig5} show the training accuracy, training loss, validation accuracy and validation loss respectively for the  model trained with EfficientNetB0 base and a fully connected final layer.It is observed that during training there is a lot of fluctuation and instability.In fact the validation curves are also not stable.
By just adding the NODE block in end before the final layer, it is seen that the training process stabilizes. \newline
Figures-\ref{fig:fig6},\ref{fig:fig7},\ref{fig:fig8},\ref{fig:fig9} depict the accuracy and loss graphs of the proposed model with the RK solver.\newline
Figure \ref{fig:fig6} depicts the training graph, where it easily attains training accuracy of about 98.5 \% in just 100 epochs.
Figure \ref{fig:fig7} shows the steady decrease in the loss of the proposed model.
Figure \ref{fig:fig8} depicts the validation curve. It is observed to be very stable in comparison to the previous model.
Figure \ref{fig:fig9} shows the corresponding validation loss.
\\
\newline
Figures-\ref{fig:fig10},\ref{fig:fig11} and \ref{fig:fig12} show the training accuracy, training loss and validation accuracy of the proposed model with the adjoint sensitivity method. Training accuracy of 99.2 \% and a validation accuracy of 85.3 \% is observed.\\ The adjoint sensitivity method provides even more stability than the Runge-Kutta Solver. It also provides better validation accuracy, and is quicker in convergence.\\
\\
Table \ref{table:table3} shows the performance of the proposed model in comparison to a model with just EfficientNetB0. Table \ref{table:table3} shows its performance on train, validation and test sets.\newline 
Table \ref{table:table2} shows the parameters set for the proposed model with the RK solver.\newline
Table \ref{table:table1} shows the number of epochs both models were trained on the corresponding time taken to train. We observe that although that the proposed model takes more time per epoch, the total time taken to converge is much better.\newline\newline
The EfficientNetB0 model and proposed model with RK solver were both developed on Keras with  Tensorflow\cite{tensorflow2015-whitepaper} backend. The proposed model with the adjoint sensititvity method was developed on PyTorch\cite{NEURIPS2019_9015}. \newline
All the above models were trained on  GTX 1080 Ti GPUs with 8GB memory.

\newpage

\newpage

\newpage

\begin{table}[ht]
\caption{Performance}
\begin{tabular}{|p{2cm}|p{2cm}|p{2cm}|p{1cm}|  }
 \hline
 \multicolumn{4}{|c|}{Performance of our model} \\
 \hline
 Model Name& Training accuracy&Validation Accuracy&Test Accuracy\\
 \hline
EfficientNetB0& 98.7\%&84.5\%&81\%\\
 \hline
Proposed Model(RK)&98.5\% &84.7\%& 81\% \\
 \hline
Proposed Model(Adjoint)&99.2\% &85.3\%&81\% \\

 \hline
\end{tabular}
\label{table:table3}
\end{table}

\begin{table}[ht]
\caption{Time Taken}
\begin{tabular}{ |p{3cm}|p{1cm}|p{1.5cm}|  }
\hline
Model Name& Epochs & Time Taken \\
\hline
EfficientNetB0 & 200 & 4.5 hours \\
 \hline
Proposed Model(RK) & 100& 3.75 hours\\
\hline
Proposed Model(Adjoint) & 160& 3.25 hours\\
 \hline
\end{tabular}
\label{table:table1}
\end{table}

\begin{table}[ht]
\caption{Parameters Set}
\begin{tabular}{|p{2.5cm}|p{2.5cm}|p{2.5cm}| }

 \hline
 Model Name& Absolute Tolerance& Relative Tolerance \\
 \hline
Proposed Model(RK) &  $10^-5$ & $10^-5$ \\

 \hline
\end{tabular}
\label{table:table2}
\end{table}

\newpage

\section{Future Work}
Using continuous depth models for transfer learning gives more freedom to choose between a trade off between accuracy and time by optimizing the tolerance values.\newline
Other solvers can be explored and examined. Exploring different methods can provide us with suitable solvers which can have better numerical stability and precision.
NODE can used for transfer learning with different network backbones to see which networks perform better.
By optimizing the model further, NODE can be a benchmark for transfer learning for most of the common datasets used in Deep Learning.
\bibliography{tlnode.bib}
\bibliographystyle{plain}

\vspace{12pt}
\color{red}
\end{document}